\documentclass[10pt,twocolumn,letterpaper]{article}

\usepackage{cvpr}
\usepackage{times}
\usepackage{epsfig}
\usepackage{amsfonts,amsmath,amsthm,amsxtra,amssymb,mathrsfs,dsfont}
\usepackage{mathtools}
\usepackage{array}
\usepackage{varwidth}
\usepackage{algorithm}
\usepackage{algpseudocode}

\usepackage[pagebackref=true,breaklinks=true,letterpaper=true,colorlinks,bookmarks=false]{hyperref}

\cvprfinalcopy 

\def\cvprPaperID{4121} 

\ifcvprfinal\fi

\title{Self-Similarity Based Time Warping}

\author{Christopher J. Tralie\\
Duke University Department of Mathematics\\
{\tt\small ctralie@alumni.princeton.edu}
}

\newcommand{\metricspace}{\ensuremath{(\mathcal{M}, d)}}
\newcommand{\ghdist}{Gromov-Hausdorff Distance}

\newcommand{\warppath}{\ensuremath{\mathcal{W}}}
\newcommand{\allwarppath}{\ensuremath{\Omega}}
\newcommand{\pStress}{\ensuremath{\mathcal{S}}}
\newcommand{\allcorresp}{\ensuremath{\Pi}}
\newcommand{\corresp}{\ensuremath{\mathcal{C}}}

\newcommand{\frechet}{Fr\'{e}chet Distance}
\newcommand{\topc}{time-ordered point cloud}

\newtheorem{theorem}{Theorem}
\newtheorem{lemma}[theorem]{Lemma}

\begin{document}


\maketitle

\begin{abstract}
In this work, we explore the problem of aligning two time-ordered point clouds which are spatially transformed and re-parameterized versions of each other.  This has a diverse array of applications such as cross modal time series synchronization (\eg MOCAP to video) and alignment of discretized curves in images.  Most other works that address this problem attempt to jointly uncover a spatial alignment and correspondences between the two point clouds, or to derive local invariants to spatial transformations such as curvature before computing correspondences.  By contrast, we sidestep spatial alignment completely by using self-similarity matrices (SSMs) as a proxy to the time-ordered point clouds, since self-similarity matrices are {\em blind to isometries} and respect global geometry.  Our algorithm, dubbed ``Isometry Blind Dynamic Time Warping'' (IBDTW), is simple and general, and we show that its associated dissimilarity measure lower bounds the L1 Gromov-Hausdorff distance between the two point sets when restricted to warping paths.  We also present a local, partial alignment extension of IBDTW based on the Smith Waterman algorithm.  This eliminates the need for tedious manual cropping of time series, which is ordinarily necessary for global alignment algorithms to function properly.
\end{abstract}

\section{Introduction / Background}
\begin{figure}[ht]
\centering
\includegraphics[width=\columnwidth]{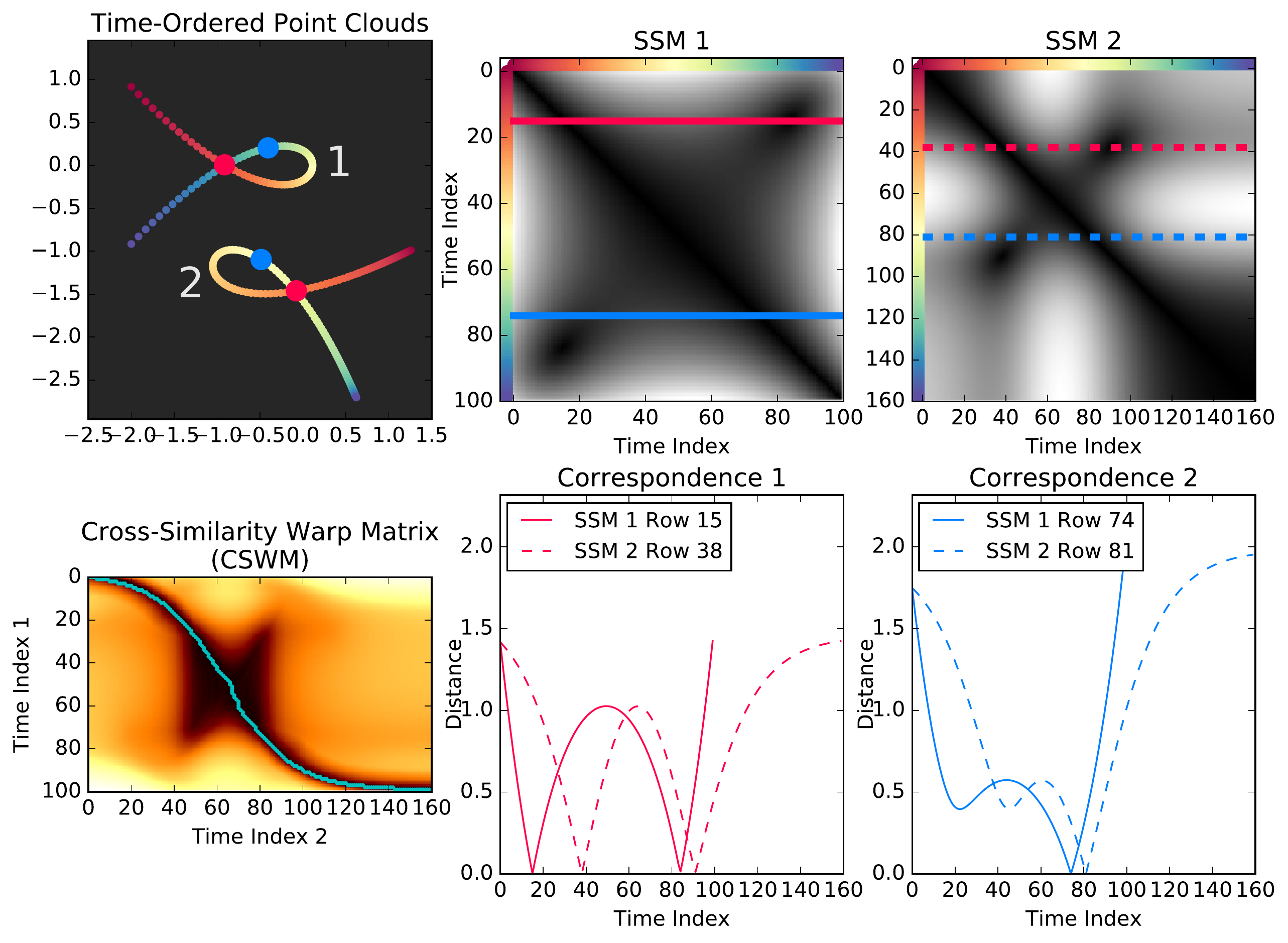}
\caption{A concept figure for our technique of aligning time-ordered point clouds which are rotated/translated/flipped and re-parameterized versions of each other.  Rows of self-similarity matrices (SSMs) of points which are in correspondence are re-parameterized versions of each other, which reduces the global alignment problem to a series of 1D time warping problems.  This observation forms the basis of our algorithm, which returns ``warping path,'' drawn in cyan in the lower left plot of this figure, that informs how to synchronize the point clouds.}
\label{fig:IntroFig}
\end{figure}

In this work, we address the problem of synchronizing sampled curves, which we refer to as ``time-ordered point clouds'' (TOPCs).  The problem of synchronizing TOPCs which trace similar trajectories but which may be parameterized differently is usually approached with the Dynamic Time Warping (DTW) algorithm \cite{sakoe1970similarity,sakoe1978dynamic}.  Since sequential data can often be translated into a sequence of vectors in some feature space, this algorithm has found widespread use in applications such as spoken word synchronization \cite{sakoe1970similarity, sakoe1978dynamic}, gesture recognition \cite{ten2007multi}, touch screen authentication \cite{de2012touch}, video contour shape sequence alignment \cite{maurel2003dynamic}, and general time series alignment \cite{berndt1994using}, to name a few of the thousands of works that use it.  The problem becomes substantially more difficult, however, when the point clouds undergo spatial transformations or dimensionality shifts in addition to re-parameterizations, which is more common {\em across} modalities.  For instance, one may want to synchronize a motion capture sequence expressed with quaternions of joints with a video of a similar motion in some feature space (see Section~\ref{sec:noneuclidean}).  There is no apparent correspondence between these spaces a priori.  This problem even arises within modalities, such as aligning gestures from different people who reside in different spatial locations.  Thus, when synchronizing sampled curves, it is important to address not only re-parameterizations, but also spatial transformations such as maps between spaces or rotations/translations/flips within the same space.

In our work, we avoid explicitly solving for spatial maps by using {\em self-similarity matrices} (SSMs).  Figure~\ref{fig:IntroFig} shows a sketch of the technique.  Even if the curves have been rotated/translated/flipped and re-parameterized, rows of the SSMs which are in correspondence are re-parameterized versions of each other.  Our technique is simple both conceptually and in implementation, it is fully unsupervised, and it is parameter free.  There are also theoretical connections between our algorithm and metric geometry, as shown in Section~\ref{sec:theory}.  We also show an extension of our basic technique to {\em partially} align time series across modalities (Section~\ref{sec:partialalignment}), and this is the first known solution to that problem.  Finally, with proper normalization (Section~\ref{sec:normalization}), these techniques can address cross-modal alignment.  We show favorable results on a number of benchmark datasets (Section~\ref{sec:globalexperiments}).

\subsection{Self-Similarity Matrices (SSMs)}

The main data structure we rely on in this work is the {\em self-similarity matrix}.  Given a space curve parameterized by the unit interval $\gamma: [0, 1] \rightarrow \metricspace$, a {\em Self-Similarity Image (SSI)} is a function $D: [0, 1] \times [0, 1] \rightarrow \mathbb{R}$ so that

\begin{equation}
D_{\gamma}(i, j) = d(\gamma(i), \gamma(j))
\end{equation}

The discretized version of an SSI corresponding to a sampled version of a curve is a {\em self-similarity matrix (SSM)}.  SSIs and SSMs are naturally {\em blind to isometries} of the underlying space curve and time-ordered point cloud, respectively; these structures remain the same if the curve/point cloud is rotated/translated/flipped.

Time-ordered SSMs have been applied to the problem of human activity recognition in video \cite{junejo2008cross}, periodicity and symmetry detection in video motion \cite{cutler2000robust}, musical audio note boundary detection \cite{foote2000automatic}, music structure understanding and segmentation \cite{bello2009SSMStructure, kaiser2010music, serra2012unsupervised, mcfee2014analyzing}, cover song identification \cite{tralie2015coversongstimbral}, and dynamical systems \cite{mcguire1997recurrence}, to name a few areas.  In this work, we study more general properties of time-ordered SSMs that make them useful for alignment.

\subsection{Warping Paths / Dynamic Time Warping}
The warping path is the basic primitive object we seek to synchronize two time-ordered point clouds.  It is a discrete version of an orientation preserving homeomorphism of the unit interval used to re-parameterize curves.  In layman's terms, it provides a way to step forward along both point clouds jointly in a continuous way without backtracking, so that they are optimally aligned over all steps.  More precisely, given two sets $X$ and $Y$, a {\em correspondence} $\corresp$ between the two sets is such that $\corresp \subset X \times Y$ and $\forall x \in X \exists y \in y$ s.t. $(x, y) \in \corresp$ and $\forall y \in Y \exists x \in X$ s.t. $(x, y) \in \corresp$.  In other words, a correspondence is a matching between two sets $X$ and $Y$ so that each element in $X$ is matched to at least one element in $Y$, and each element of $Y$ is matched to at least one element in $X$.  Let $X$ and $Y$ be two sets whose elements are adorned with a time order: $X = \{x_1, x_2, ..., x_M\}$ and $Y = \{y_1, y_2, ..., y_N\}$.  A {\em warping path}, between $X$ and $Y$ is a correspondence $\warppath$ which can be put into the sequence $\warppath = (c_1, c_2, ..., c_K)$ satisfying the following properties
\begin{itemize}
\item {\em Monotonicity}: If $(x_i, y_j) \in \warppath$, then $(x_k, y_l) \notin \warppath$ for $k < i$, $l > j$
\item {\em Boundary Conditions}: $(x_1, y_1), (x_M, y_N) \in \warppath$
\item {\em Continuity}: $c_i - c_{i-1} \in \{ (0, 1), (1, 0), (1, 1) \}$
\end{itemize}

Now suppose there are two time-ordered point clouds $X$ and $Y$ which both live in the same metric space $\metricspace$.  The {\em Dynamic Time Warping (DTW) Dissimilarity}\cite{sakoe1970similarity,sakoe1978dynamic}\footnote{Note that DTW is not a metric, as it fails to satisfy the triangle inequality.  For an example, see \cite{muller2007information} section 4.1} between $X$ and $Y$ is

\[ \text{DTW}(X, Y) = \min_{\warppath \in \allwarppath}  \sum_{(i, j) \in \warppath} d(x_i, y_j) \]

where $\allwarppath$ is the set of all valid warping paths between $X$ and $Y$.  DTW satisfies the following {\em subsequence relation}

\begin{equation}
DTW_{ij} =  \left\{ d(x_i, y_j) + \min \begin{array}{c} DTW_{i-1, j-1} \\ DTW_{i-1, j} \\ DTW_{i, j-1} \end{array}  \right\}
\end{equation}

where $DTW_{ij}$ is the DTW dissimilarity between $\{x_1, x_2, ..., x_i\}$ and $\{y_1, y_2, ..., y_j\}$.  This makes it possible to solve DTW with a dynamic programming algorithm which takes $O(MN)$ time.  This algorithm computes the cost of the shortest path from the upper left to the lower right of the ``cross-similarity matrix'' (CSM), or the $M \times N$ matrix holding all distances $d_{ij}$ between $X_i$ and $Y_j$.  A (not necessarily unique) shortest path realizing this distance is a warping path which can be used to align the two time series.

\subsection{DTW After Spatial Transformations}

In addition to synchronizing curves in the same ambient space which are approximately re-parameterizations of each other, there has also been some recent work on the more difficult problem of matching curves which live in different ambient spaces or which live in the same space but which may differ by a spatial transformation in addition to re-parameterization.  One objective for spatial alignment is the optimal {\em rigid transformation} taking one set of points to another.  More precisely, given two Euclidean point clouds $X, Y \in \mathbb{R}^d$, each with $N$ points which are assumed to be in correspondence, the {\em Procrustes distance} \cite{whaba1965least, kabsch1976solution} is

\begin{equation}
d_P(X, Y) = \min_{R_x, R_y, t_x, t_y} \sum_{i = 1}^N ||R_x (x_i - t_x) - R_y(y_i - t_y)||_2^2
\end{equation}

One issue with Procrustes is that not only do $X$ and $Y$ have to have the same number of points, but the correspondences must be known a priori.  Often in practice, neither of these assumptions are true.  To deal with this, one can use the ``Iterative Closest Points'' (ICP) algorithm \cite{besl1992method, chen1992object}, which switches back and forth between finding correspondences with nearest neighbors and solving the Procrustes problem.  The authors of \cite{Ying2016DTW} use a modified version of ICP, replacing the nearest neighbor correspondence step with DTW.  This ensures that the time order will be respected, which is not guaranteed with nearest neighbors only\footnote{This analogous to the difference between the \frechet{} \cite{alt1995computing} and the Hausdorff Distance between curves.}.

There are also techniques which use canonical correlation analysis (CCA) instead of Procrustes analysis.  Given two point clouds with $N$ points represented by matrices $X \in \mathbb{R}^{d_1 \times N}$ and $Y \in \mathbb{R}^{d_2 \times N}$, assumed to be in correspondence, CCA is defined as

\begin{equation}
d_{\text{CCA}} = \min_{V_x \in \mathbb{R}^{d_x \times b}, V_y} ||V_x^T X - V_y^T Y||_F^2
\end{equation}

for some chosen constant $b \leq \min(d_1, d_2)$, s.t. $V_x^T X X^T V_x = V_y^T Y Y^T V_y = I_b$.  This is better suited to cross modal applications where scaling is involved.  Like Procrustes, this assumes that the correspondences are known a priori.  To find the correspondences, the authors in \cite{zhou2009canonical} take the same iterative approach as that authors in \cite{Ying2016DTW} did with ICP, but they alternate back and forth between DTW and CCA instead of DTW and Procrustes.  An updated version of this algorithm known as ``generalized time warping'' (GTW) \cite{zhou2012generalized, zhou2016generalized} was developed which aligns multiple sequences using a single optimization objective where the spatial alignment and time warping are coupled.  Finally, a recent work in \cite{trigeorgis2017deep, trigeorgis2016deep} takes a similar approach, but it replaces CCA by learning features in the projection stage with a deep neural network.  Like all supervised learning approaches, however, this method requires training data with known correspondences.  Furthermore, all of the techniques we have mentioned so far require a good initial guess to converge to a globally optimal solution.

As an alternative to solving for a spatial alignment explicitly, many works perform time warping on a surrogate function which is invariant to isometries of the input.  A popular choice is to numerically estimate curvature \cite{keogh2001derivative, sebastian2003aligning, frenkel2003curve}.  Some works use the triangle area between triples of points as an invariant \cite{alajlan2007shape}, and some use the turning angle of the curve \cite{cohen1997partial}, which is related to curvature.  These techniques can suffer from numerical difficulties when estimating the invariants.  Also, most of the invariants are local, so small differences can cause the curves ``drift'' over time (\eg a U is similar to a 6 with local curvature \cite{sebastian2003aligning}), though using integrated curvature \cite{cui2009curve} can ameliorate this.

Beyond spatial alignment and invariants, the authors of \cite{yamada2015cross} address more general case of cross-domain object matching (CDOM) with general correspondences and address warping paths as a special case.  However, their problem reduces to the quadratic assignment problem, which is NP-hard, and their iterative approximation requires a good initial guess. The authors of \cite{vejdemo2015cohomological} address a special case where curves form closed loops, using cohomology to find maps from point clouds to the circle, where they are synchronized, but this only works for periodic time series.  The authors of \cite{gong2011dynamic} jointly align curves on manifolds, which is effective but requires learning the manifolds.  Perhaps the most similar to our approach is the action recognition work of \cite{junejo2011view}, from which we drew much inspiration, which applies DTW to small patches of SSMs to align time warped actions from different camera views.  However, they only use elements near the diagonal of SSMs, and their scheme does not extend across modalities.

\section{Isometry Blind Dynamic Time Warping}

\begin{figure}
\includegraphics[width=\columnwidth]{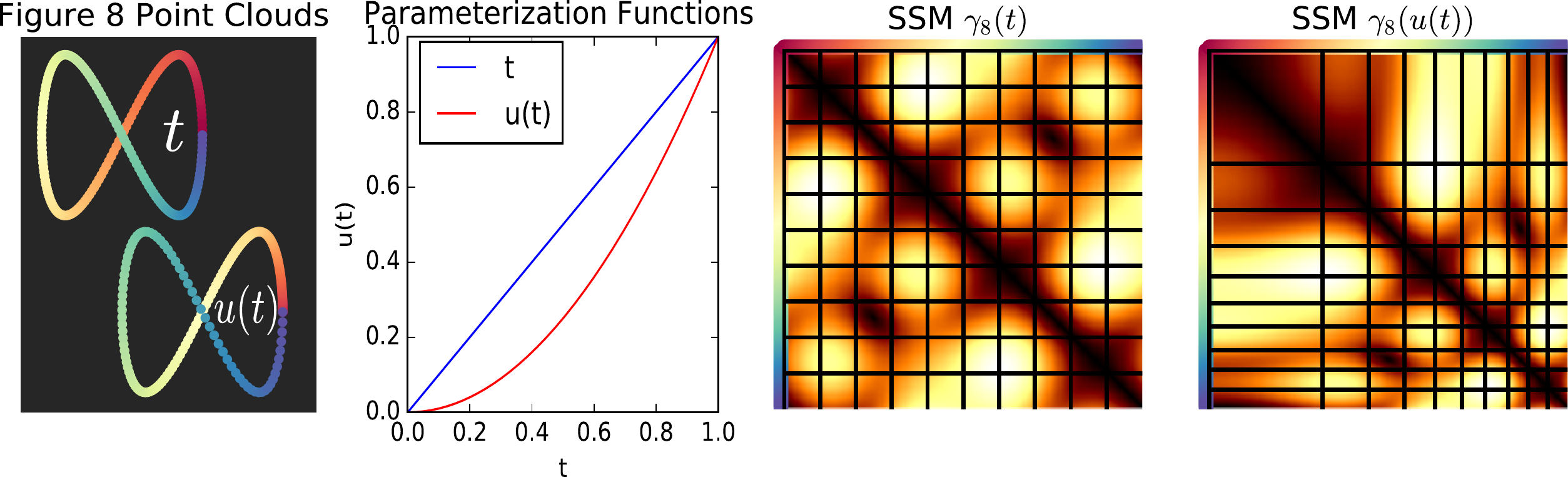}
\caption{Self-similarity images of different parameterizations of a Figure 8. Rectangles in one image map to rectangles in the other image, and lines in one image map to lines in the other image.}
\label{fig:Figure8Reparameterizations}
\end{figure}

Most of the approaches we reviewed to align time series which have undergone linear transformations try to explicitly factor out those transformations before doing an alignment, but this is not necessary if we build our algorithm on top a self-similarity matrix between two point clouds, which is already blind to isometries.  To set the stage for our algorithms, we first study the maps that are induced between self-similarity images by re-parameterization functions, which will help in the algorithm design.  For example, take the figure 8 curve,$\gamma_8(t) = (\cos (2 \pi t), sin (4 \pi t))$.  The bottom left of Figure~\ref{fig:Figure8Reparameterizations} shows the SSM of a linearly parameterized sampled version of this curve, while the bottom right of Figure~\ref{fig:Figure8Reparameterizations} shows the SSM corresponding to a re-parameterized sampled version.  Maps between the domains of the SSMs shown are always rectangles, and they are independent of underlying curve being parameterized (they only depend on the relationship between two parameterizations).  To see this, start with a space curve $\gamma: [0, 1] \rightarrow \mathbb{R}^d$ and its resulting self-similarity image $D_{\gamma}$.  Given a homeomorphism $h: [0, 1] \rightarrow [0, 1]$, which yields a space curve $\gamma_h: [0, 1] \rightarrow \mathbb{R}^d$ and a corresponding self-similarity image $D_{\gamma_h}$, there is an induced homeomorphism, $h \times h$ from the square to itself between the two domains of $D_{\gamma}$ and $D_{\gamma_h}$; that is, $ D_{\gamma_h} = D_{\gamma}(h(s), h(t))$.  If we fix a correspondence $s \iff u = h(s)$, then this shows that row $h(s)$ of $D_{\gamma}$ is a 1D re-parameterization of row $s$ of $D_{\gamma_h}$, making rigorous the observation in Figure~\ref{fig:IntroFig}.  Note that for a discrete version of these maps between \topc s, one can replace the homeomorphism $h$ with a warping path \warppath{}, and the relationships are otherwise the same.

\subsection{IBDTW Algorithm}

We now have the prerequisites necessary to define our main algorithm.  The idea is quite simple.  Based on our observations and Figure~\ref{fig:IntroFig} and Figure~\ref{fig:Figure8Reparameterizations}, if we know that point $i$ in a time-ordered point cloud (TOPC) $X$ is in correspondence with a point $j$ in TOPC Y, then we should match the $i^{\text{th}}$ row of X's SSM to the $j^\text{th}$ row of Y's SSM under the $L_1$ distance, enforcing the constraint that $(i, j) \in \warppath$.  However, since it is unknown a priori which rows should be in correspondence, we try every row $i$ of SSM A against every row $j$ in SSM B, and we create a {\em cross-similarity time warping matrix (CSWM)} $C$ so that $C_{ij}$ contains $L_1$ DTW between row $i$ of SSMA and row $j$ of SSMB, {\em constrained to warping paths which include $(i, j)$}.  To enforce that $(i, j)$ be in the optimal warping path,  we exploit the boundary condition property of DTW by running the original DTW algorithm twice: once between $SSMAi_{1:i}, SSMBj_{1:j}$ and once between $SSMAi_{i:M}$ and $SSMBj_{j:N}$, summing the costs.   After doing this $\forall i, j$, apply the ordinary DTW algorithm to $C$.  Algorithm~\ref{alg:ibdtw} summarizes this process.  Note that a serial implementation of this algorithm takes $O(M^2N^2)$ time, since a 1D DTW is computed for every row pair.  To mitigate this, we implement a linear systolic array\cite{yu2005smith} version of DTW in CUDA.  With unlimited parallel processors, this reduces computation to $O(M+N)$.  In practice, we witness a 30x speedup between point clouds with hundreds of samples.

\begin{figure}
\centering
\includegraphics[width=\columnwidth]{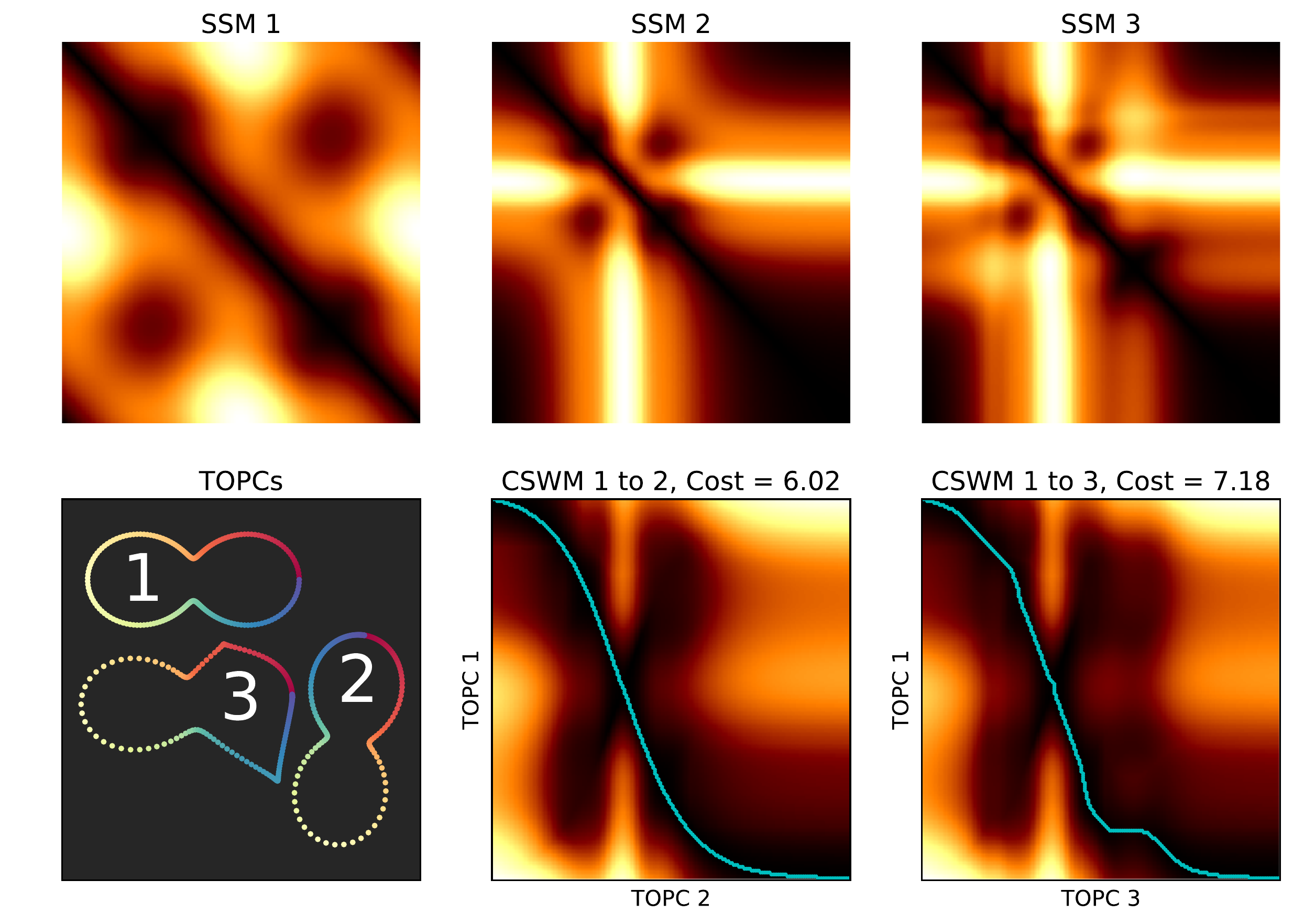}
\caption{An example of IBDTW between 3 different samplings of a pinched ellipse.  The optimal warping path found by Algorithm~\ref{alg:ibdtw} is drawn in cyan on top of the CSWM in each case.  Based on this, points which are in correspondence are drawn with the same color in the lower left figure.  Though time-ordered point cloud 2 has more points towards the beginning and fewer points towards the end than time-ordered point cloud 1, correct regions are put into correspondence with each other.  Furthermore, in addition to being parameterized this way, the time-ordered point could 3 is also distorted geometrically, but the correspondences are still reasonable.}
\label{fig:IBDTWExample}
\end{figure}

Figure~\ref{fig:IBDTWExample} shows an example of this algorithm on two rotated/translated/re-parameterized \topc s in $\mathbb{R}^2$ (point clouds 1 and 2).  As the colors show, IBDTW puts the points into correspondence correctly even without first spatially aligning them.  We also show alignment to a third time-ordered point cloud, which is metrically distorted in addition to being rotated/translated/re-parameterized.  The returned warping degrades gracefully.  We will explore this more rigorously in Section~\ref{sec:curvealignment}.

\begin{algorithm}[t]
  \caption{Isometry Blind Dynamic Time Warping}\label{alg:ibdtw}
  \begin{algorithmic}[1]
    \Procedure{IBDTW}{$X$, $Y$, $d_X$, $d_Y$}
    \State $\triangleright$ TOPCs $X$ and $Y$ with $M$ and $N$ points, metrics $d_X$ and $d_Y$, respectively
    %
    \sbox0{$\vcenter{\hbox{$\begin{array}{|c|c|c|c|c|}
      \hline
      0 & \infty & \infty & \ldots & \infty \\ \hline
      \infty & 0 & 0 & \ldots & 0 \\ \hline
      \vdots & \vdots & \vdots & \ldots & \vdots \\ \hline
      \infty & 0 & 0 & \ldots & 0 \\ \hline
    \end{array}$}}$}%
    \State $\triangleright$ Initialize cross-similarity warp matrix (CSWM) \\
    C $\leftarrow
      \underbrace{\vrule width0pt depth \dimexpr\dp0 + .3ex\relax\copy0}_{N}%
      \left.\kern-\nulldelimiterspace
        \vphantom{\copy0}
      \right\rbrace \scriptstyle M
    $ \label{algline:C}
    \For{$i = 1:M$}
        \State $\triangleright i^{\text{th}}$ row of $d_X$
        \State $A \gets [d_X(x_i, x_1), d_X(x_i, x_2), \dots, d_X(x_i, x_M)]$
        \For{$j = 1:N$}
            \State $\triangleright j^{\text{th}}$ row of $d_Y$
            \State $B \gets [d_Y(y_j, y_1), d_Y(y_j, y_2), \dots, d_Y(y_j, y_N)]$
            \State $C_{ij} \gets \text{ConstrainedDTW(A, B, $L_1$, $i$, $j$)}$\label{algline:dtwrec}
            \label{line:innerloopdtw}
        \EndFor
    \EndFor
    \State $\triangleright$ Use the CSWM $C$ in ordinary DTW
    \State $D \gets \frac{1}{2}$DTW$(X, Y, C)$\label{algline:dtwouter}

    \Return $(D, C)$
    \Comment{Return the cost and the CSWM}
    \EndProcedure
  \end{algorithmic}
\end{algorithm}

\subsection{Analysis: Lower Bounding L1 Metric Stress}
\label{sec:theory}
IBDTW can be put into the {\em \ghdist{}} framework, which describes how to ``embed'' one metric space into another.  More formally, given two discrete metric spaces $(X, d_X)$ and $(Y, d_Y)$, and a correspondence $\corresp$ between $X$ and $Y$, the {\em $p$-stress} is defined as

\begin{equation}
\pStress_p(X, Y, \corresp) = \left( \sum_{(x, y), (x', y') \in \corresp} (d_X(x, x') - d_Y(y, y'))^p \right)^{1/p}
\end{equation}

Intuitively, the $p$-stress measures how much one has to stretch one metric space when moving it to another.  The \ghdist{} $d_{\text{GH}}$ uses $S_\infty$ specifically:

\begin{equation}
\label{eq:gromovhausdorff}
d_{\text{GH}}(X, Y) = \frac{1}{2} \inf_{\corresp \in \allcorresp} \pStress_{\infty}(X, Y, \corresp)
\end{equation}
where \allcorresp{} is the set of all correspondences between $X$ and $Y$.  In other words, the \ghdist{} measures the smallest possible {\em distortion} between a pair of points over all possible embeddings of one metric space into another.  Unfortunately, the \ghdist{} is NP-complete, but we can still connect Algorithm~\ref{alg:ibdtw} to the \ghdist{} via the following lemma:

\begin{lemma}
The cost returned by Algorithm~\ref{alg:ibdtw} lower bounds $\pStress_1(X, Y, \warppath)$, or the 1-stress {\em restricted to warping paths}.
\end{lemma}

Proof: Note that the optimal IBDTW warping path $\warppath^*$ has the following cost $c(\warppath^*)$

\begin{equation}
c(\warppath^*) = \sum_{(x_i, y_j), (x', y') \in \warppath^*} |d_X(x_i, x') - d_Y(y_j, y')|
\end{equation}

which can be rewritten as

\begin{equation}
c(\warppath^*) = \frac{1}{2} \sum_{(x_i, y_j) \in \warppath^*} \sum_{(x', y') \in \warppath^*} |d_X(x_i, x'), - d_Y(y_j, y')|
\end{equation}

since if $(x', y') = (x_i, y_j)$ then the cost is zero, all other terms counted twice.  Now fix an $x_i$ and $y_j$.  Then the sum of the terms of the form $|d_X(x_i, x') - d_Y(y_j, y')|$ is simply the $L_1$ warping distance between 1D time series which are the $i^{\text{th}}$ row of $d_X$, $d_X[i, :]$ and the $j^\text{th}$ row of $d_Y$, $d_Y[j, :]$ under the warping $\warppath^*$.  Also, the DTW Distance between $d_X[i, :]$ and $d_Y[j, :]$ is at most the $L_1$ warping distance under $\warppath^*$, and is potentially lower since we are computing them greedily only between $x_i$ and $y_j$, ignoring all other constraints.  Hence, the sum of the terms $|d_X(x_i, x') - d_Y(y_j, y')|$ is lower bounded by Line~\ref{algline:dtwrec} in Algorithm~\ref{alg:ibdtw}. $\blacksquare$

For a more direct analogy with DTW, Algorithm~\ref{alg:ibdtw} was designed to lower bound the 1-stress restricted to warping paths.  We note that a similar technique could be used to lower bound the \ghdist{} restricted to warping paths by replacing constrained DTW in the inner loop in Line~\ref{line:innerloopdtw} with a constrained version of the discrete \frechet{} \cite{eiter1994computing} to find the maximum distortion induced by putting two points in correspondence.  In this work, however, we stick to the 1-stress, since it gives a more informative overall picture of the full metric space.

\section{Isometry Blind Partial Time Warping}
\label{sec:partialalignment}

One of the drawbacks of IBDTW is that it requires a global alignment.  However, if the sequences only partially overlap, forcing a global alignment leads to poor result, unless manual cropping is done to ensure that sequences start and end at the same place \cite{zhou2009canonical}.  To automate cropping, we design an isometry blind time warping algorithm that can do partial alignment.  This algorithm is like IBDTW, except DTW is replaced with the ``Smith Waterman'' algorithm \cite{smith1981identification,waterman1995introduction}, which seeks the best contiguous {\em subsequences} in each time series which match each other \footnote{This algorithm was originally developed for gene sequence alignment, but it has been adapted to multimedia problems such as music alignment \cite{serra2008chroma} and video copyright infringement detection \cite{bronstein2010video}.}. Unlike dynamic time warping, Smith Waterman seeks to {\em maximize} an alignment {\em score}, and the alignment does not have to start on the first element of each sequence or end on the last element on each sequence.  To solve this, the exact same dynamic programming algorithm is used, except there is one extra ``restart'' condition if a local alignment has become sufficiently poor.  The recurrence is

\begin{equation}
SW_{ij} = \max \left\{ \begin{array}{c} SW_{i-1, j-1} + m(x_i, y_j) \\ SW_{i-1, j} + g \\ SW_{i, j-1} + g \\ 0 \end{array}  \right\}
\end{equation}

where $m(x_i, y_j)$ is a matching score between points $x_i$ and $y_j$, which is positive for a match and negative for a mismatch, and $g$ is a gap penalty.  

\begin{figure}
\centering
\includegraphics[width=\columnwidth]{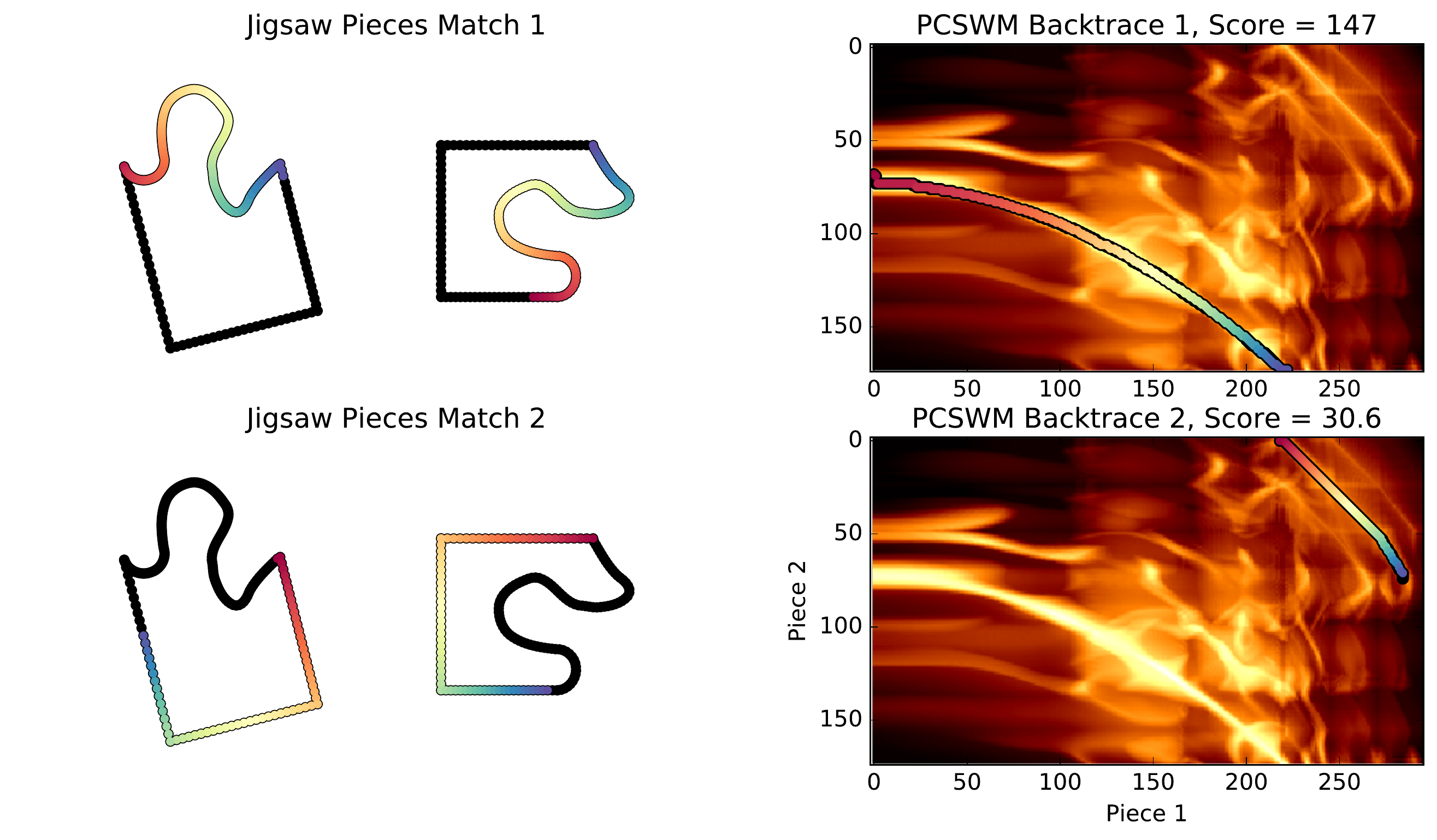}
\caption{Partial alignment with IBPTW on jigsaw puzzle pieces.  The middle row shows the optimal partial alignment.  The bottom row shows a locally optimally partial alignment with a lower score. Please refer to color version of this figure for full detail.}
\label{fig:JigsawExample}
\end{figure}

Like DTW, we may modify Smith Waterman to return the best subsequence constrained to match the $i^{\text{th}}$ point in the first sequence to the $j^{\text{th}}$ point in the second sequence by runing Smith Waterman between $\{X_1, X_2, ..., X_i\}$ and ${Y_1, Y_2, ..., Y_j}$, and then again between the reversed sequences $\{X_M, X_{M-1}, X_{M-2}, ..., X_i\}$ and $\{Y_N, Y_{N-1}, Y_{N-2}, ..., Y_1\}$.  Then, the {\em Isometry Blind Partial Time Warping (IBPTW)} algorithm is exactly like Algorithm~\ref{alg:ibdtw}, except we replace Line~\ref{algline:dtwrec} with constrained Smith Waterman, and we replace Line~\ref{algline:dtwouter} with unconstrained Smith Waterman.  We refer to the matrix $C$ (Line~\ref{algline:C}) as the ``partial cross-similarity warp matrix (PCSWM).''  In practice, we define $m_1(a, b) = \exp(-|d(a, b)|/\sigma) - 0.6$ for Line~\ref{algline:dtwrec}.  If $d(a, b)$ is the L1 distance between elements of two histogram normalized SSM rows, then it ranges between $0$ and $1$.  Thus, there is a positive matching score of up to 0.4 for the most similar SSM values and a negative matching score of -0.6 between the most dissimilar values.  We also choose a gap penalty of -0.4 to promote diagonal matches.  Otherwise, the warping path maximizing the alignment score contains longer horizontal and vertical lines, leading to undesirable pauses of one time series with respect to the other.  For the outer loop (Line~\ref{algline:dtwouter}), we use
\[ m_2(a_i, b_j) = \frac{(S_{ij} - \text{md}(S))}{\max(|S-\text{md}(S))|} \]
where md is the median operation.  This will give a high score up to $\leq 1$ to row pairs which have a high subsequence score in common, and a low score $\geq -1$ to rows which do not have a good subsequence.  Figure~\ref{fig:JigsawExample} shows an example of this algorithm on two jigsaw puzzle pieces which should fit together with $m_1$ and $m_2$ defined as above, $\sigma = 0.09$, and $g_1, g_2 = -0.4$.  The longest subsequence is indeed along the cutouts where they match together.  It is also possible to backtrace from anywhere in the PCSWM to find other subsequences which match in common, so Figure~\ref{fig:JigsawExample} also shows an example of suboptimal but good local alignment.

\section{Cross-Modal Histogram Normalization}
\label{sec:normalization}

\begin{figure}
\centering
\includegraphics[width=\columnwidth]{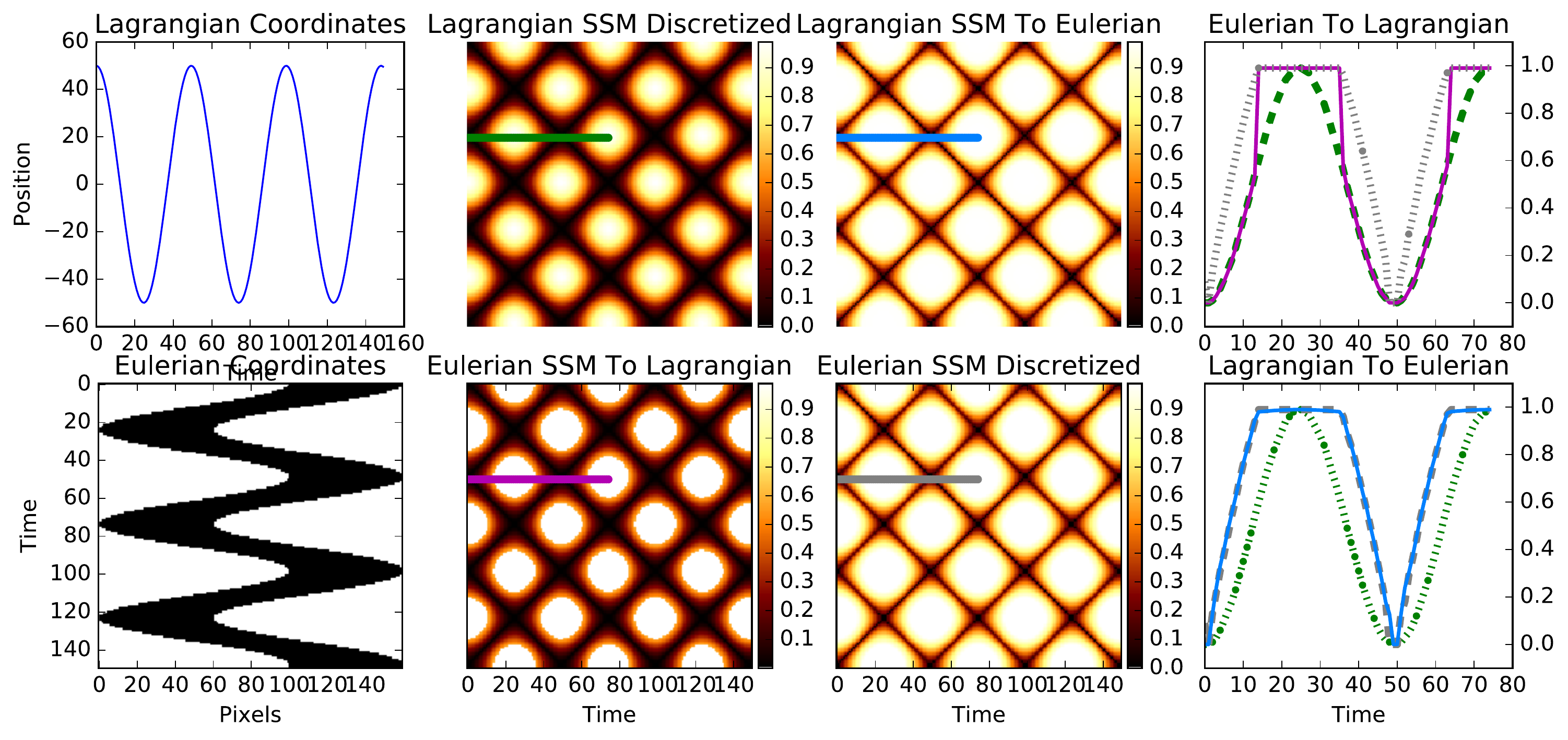}
\caption{An example of matching the SSM of an oscillating line segment captured with Lagrangian coordinates to an SSM of the oscillating line segment captured with Eulerian coordinates, and vice versa.  The right column shows an example of a row from each of the matrices in different cases.  The stipple line pattern shows the original row, the line segment shows the corresponding row from the SSM with the target distribution, and the solid row shows the remapped version of the original row.  In this case, it is easier to remap the Lagrangian coordinates to Eulerian coordinates, though both remappings are closer to the target than the original.}
\label{fig:normalization}
\end{figure}

The schemes we have presented work well for point clouds sampled from isometric curves, but the isometry assumption does not usually hold in cross-modal applications.  Not only can the scales be drastically different between modalities, but it is unlikely that a uniform re-scaling will fix the problem.  For instance, consider a 1D oscillating bar of length $B$ oscillating sinusoidally over the interval $[0, A]$ with a period of $T$.  Its center position is measured as $c_t = (A/2) + (A/2)\cos(2 \pi t / T)$. This type of measurement, which follows the object in question, is in {\em Lagrangian coordinates}.  By contrast, suppose we take a 1D video of the bar with $A$ pixels, where each pixel in each frame measures occupancy by the bar at that frame in the video.  Then pixel $i$ in video $X_t$  is parameterized by time as

\begin{equation}
X_t[i] = \left\{ \begin{array}{cc} 1 & |c_t + B/2 - i| < B/2 \\ 0 & \text{otherwise} \end{array} \right\}
\end{equation}

These pixel by pixel measurements at fixed positions are referred to as {\em Eulerian coordinates}.  Let the Lagrangian SSM $D_1$ be the 1D metric between two different centers, $D_1[s, t] = |c_s - c_t|$, and let the Eulerian SSM $D_2$ be the Euclidean metric $D_2[s, t] = ||X_s - X_t||_2$ between each frame of the video.  Although they are measuring the same process, the SSMs have a locally different character.  Rows of $D_1$ are perfect sinusoids, while rows of $D_2$ are more like square waves, since there are sharp transitions from foreground to background in Eulerian coordinates.  Figure~\ref{fig:normalization} summarizes all of this visually.

To address this kind of local rescaling between an SSM $D_1$ and an SSM $D_2$, we first divide each SSM by its respective max, and we quantize each to $L$ levels evenly spaced in $[0, 1]$.  We then apply a monotonic, one-to-one map $f$ to each pixel in $D_1$ so that the CDF of $D_1$ approximately matches the CDF of $D_2$ (see, \eg, \cite{gonzalez1992digital} ch. 3.3).  Note that this process can be done from $D_1$ to $D_2$ or from $D_2$ to $D_1$, as shown in Figure~\ref{fig:normalization}.  Since this process is not necessarily symmetric, we perform both sets of normalizations, and we choose the one which yields a better alignment score.

\section{Experimental Results}
\label{sec:globalexperiments}

In this section, we will quantitatively compare the IBDTW algorithm for global alignment with several other techniques in the literature, including ordinary dynamic time warping (DTW), derivative dynamic time warping (DDTW) \cite{keogh2001derivative} (a curvature-based version), canonical time warping (CTW)\cite{zhou2009canonical}, Generalized Time Warping (GTW)\cite{zhou2012generalized, zhou2016generalized}, and Iterative Motion Warping (IMW)\cite{hsu2005style} (a simpler version of CTW which is restricted to the same space).  We use code from \cite{zhou2009canonical} and \cite{zhou2012generalized, zhou2016generalized} to compute all of these alignments\footnote{\url{http://www.f-zhou.com/ta_code.html}}.  We use the default parameters provided in this code, and, as in \cite{zhou2009canonical} and \cite{zhou2012generalized, zhou2016generalized}, we use the results of DTW to initialize CTW and GTW.  In all of our experiments, we show results both from IBDTW and IBDTW after SSM rank normalization, which we refer to as ``IBDTWN.''  When a ground truth warping path exists, we report the alignment error as in \cite{zhou2016generalized} and \cite{trigeorgis2017deep}.  Given a warping path $\warppath = (x_1, x_2, ..., x_M)$ and a ground truth path $\warppath_{GT} = (y_1, y_2, ..., y_N)$, the alignment error is

\begin{equation}
\frac{1}{M+N}\left(  \sum_{i=1}^M \min_{j=1}^N ||x_i - y_j||_2 + \sum_{j=1}^N \min_{i=1}^M ||x_i - y_j||_2  \right)
\end{equation}

 which is (roughly) the average number of samples by which $\warppath$ is shifted from $\warppath_{GT}$ at any point in time.

\subsection{Curve Alignment}
\label{sec:curvealignment}

\begin{figure}
\centering
\includegraphics[width=\columnwidth]{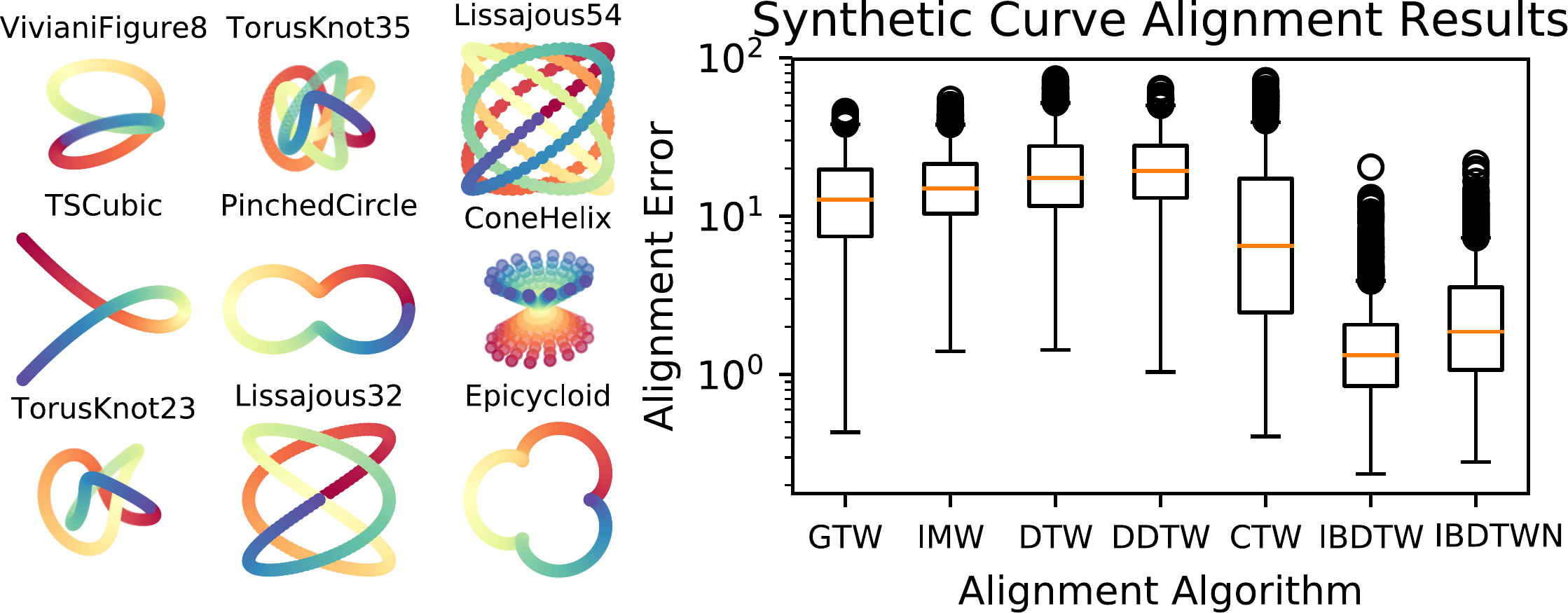}
\caption{Comparisons of alignment error distributions for different techniques on synthetic rotated/translated/re-paramterized/distorted 2D/3D curves drawn from the classes shown on the left.  Log plot shown for contrast since IBDTW performs so well relative to other methods.}
\label{fig:SyntheticExperiments}
\end{figure}

\begin{figure}
\centering
\includegraphics[width=0.8\columnwidth]{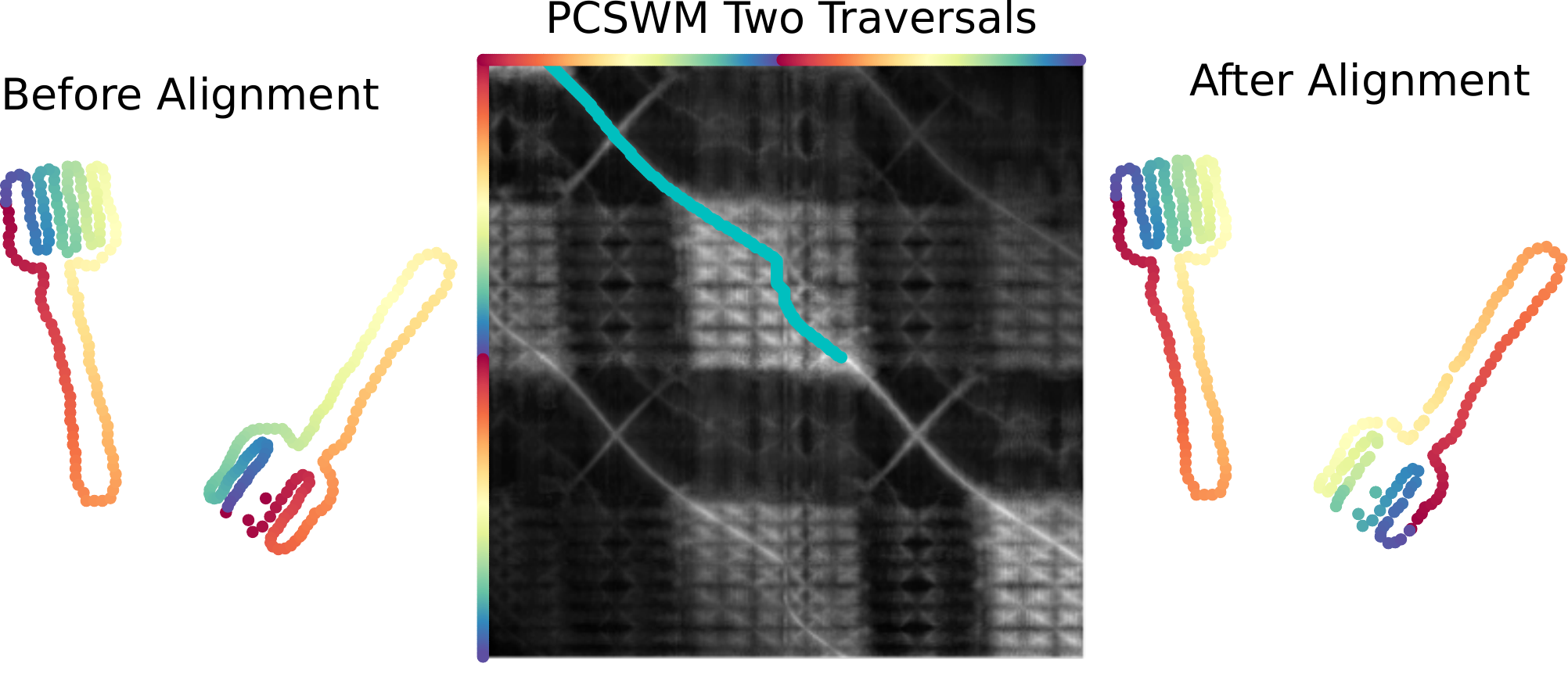}
\caption{Two closed loops in the shape of a fork which have been rotated/translated/re-parameterized, in addition to starting at different points.  The left plot shows the forks before alignment.  The center plot shows the PCSWM resulting from aligning both point clouds each repeating themselves twice, as shown by the colors along the rows (left fork) and columns (right fork), which correspond to the colors in the left plot.  The optimal partial warping path truncated to the first repetition of the left fork is superimposed in cyan.  The forks in the right plot are put into correct correspondence by this truncated partial warping path.}
\label{fig:ForkLoop}
\end{figure}

\begin{figure}
\centering
\includegraphics[width=\columnwidth]{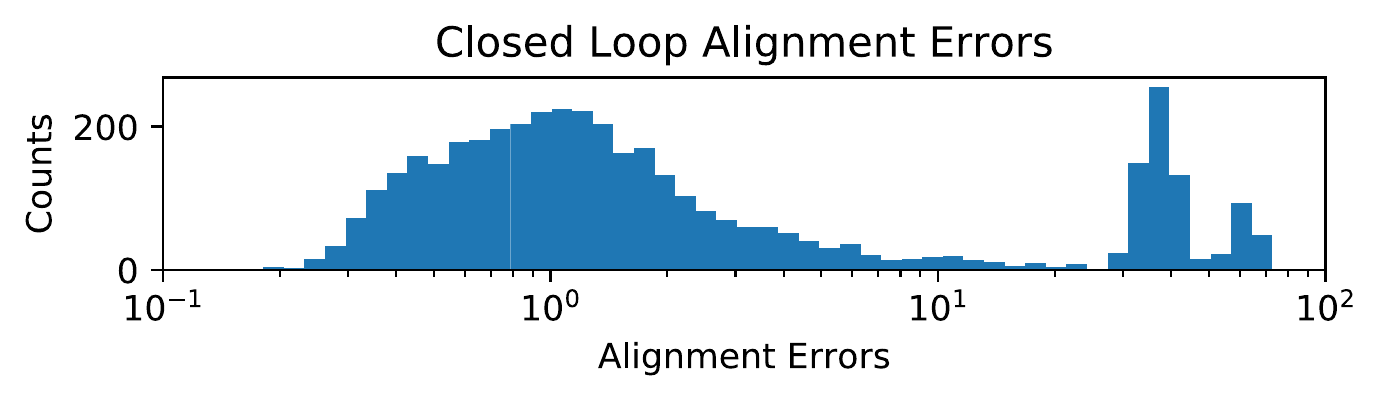}
\caption{Distribution of IBPTW alignment errors for circularly shifted/warped/distorted MPEG-7 loops.}
\label{fig:LoopsHist}
\end{figure}

We first perform an experiment aligning a series of rotated/translated/flipped and re-parameterized sampled curves.  As in \cite{zhou2016generalized}, we re-parameterize the curves with random convex combinations of polynomial, logarithmic, exponential, and hyperbolic tangent functions.  To distort the curves, we move random control points in random directions after spatial transformation and re-parameterization.  Let $X$ be the TOPC before transformation/re-paramterization/distortion and let $Y$ be the curve afterwards.  The average ratio of $d_{GH}(X, Y) / \text{diam}(X)$, where diam is the ``diameter'' of $X$ (the maximum inter point distance), is 0.18.  Figure~\ref{fig:SyntheticExperiments} shows the results.  IBDTW performs the best, while doing the normalization for IBDTWN only degrades the results slightly.

We also showcase our IBPTW algorithm by aligning 2D loops, or curves $\gamma: [0, 1] \rightarrow \mathbb{R}^2$ so that $\gamma(0) = \gamma(1)$, which is useful in recognizing boundaries of foreground objects in video.  Note that a geometrically equivalent loop can be parameterized starting at a different point in the interior of the first loop: $\gamma'(t) = \gamma(t - \tau (\text{mod} 1))$.  Also, it is possible that this loop is parameterized differently: $\gamma'_h(t) = \gamma'(h(t))$ for some orientation preserving homeomorphism $h: [0, 1] \rightarrow [0, 1]$, and $\gamma'_h(t)$ may also be transformed spatially.  To demonstrate how our algorithm is able to align such curves, we use examples from the MPEG-7 dataset of 2D contours \cite{latecki2002shape}.  Given a point cloud $A$ and a point cloud $B$, we partially align the concatenated point clouds $AA$ and $BB$.  If $A$ starts $T$ samples later than $B$, then there will be a partial warping path which starts at sample $1$ in $AA$ and sample $T$ in $BB$.  Figure~\ref{fig:ForkLoop} shows an example where two forks are successfully put into correspondence this way.  Figure~\ref{fig:LoopsHist} shows a histogram of alignment errors for distorted/circularly shifted/re-parameterized curves over 7 classes from the MPEG-7 dataset, using $\sigma = 0.01$ and $m_1, m_2 = -0.4$, and with a mean $d_{GH}/\text{diam} = 0.11$.  IBPTW returns excellent alignments for most loops, though there are a cluster of outliers that occur due to (near) symmetries of some loops.

\subsection{Weizmann Walking Videos}
\label{sec:weizmannibdtw}

\begin{figure}
\label{fig:ConstrainedDTWExample}
\includegraphics[width=\columnwidth]{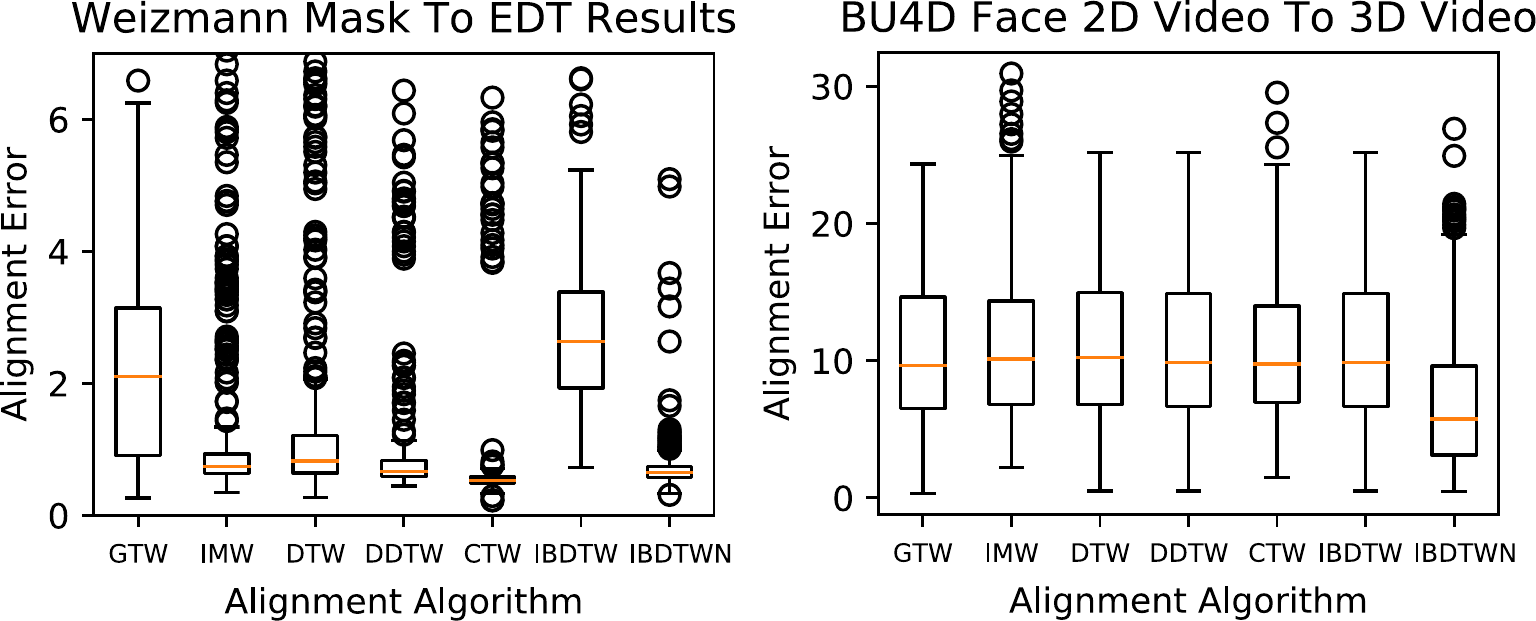}
\caption{Comparison of IBDTW with different time warping techniques on walking videos from the Weizmann action dataset\cite{weizmannPami07} (left) and on 2D/3D facial expression videos from the BU4D dataset \cite{yin2008high} (right).}
\label{fig:WeizmannBU}
\end{figure}

For our first cross modal experiment, we align 4 videos of people walking from the Weizmann dataset \cite{weizmannPami07} cropped to 4 walking cycles each.  As in \cite{zhou2009canonical}, we use different features between each pair of videos we align.  On one video, we use the binary mask of the foreground object of the person walking, where every pixel is a dimension, and every frame is a point in the TOPC.  On the second video, we use the Euclidean distance transform (EDT) \cite{maurer2003linear}.  To assess performance rigorously, we create a more controlled experiment where the second video is the same as the first video after a time warp and applying EDT, so that we have access to the ground truth.  Figure~\ref{fig:WeizmannBU} shows the results.  CTW performs slightly better than IBDTWN, but not appreciably.  Furthermore, IBTW without normalization has the worst average alignment error, demonstrating how normalization is needed in cross-modal applications.

\subsection{BU4D 2D/3D Facial Expressions}
To show a more difficult cross-modal application, we also synchronize expressions drawn from the BU4D face dataset \cite{yin2008high}, which includes RGB videos of people making different facial expressions and a 3D triangle mesh corresponding to each frame.  For our RGB features, we simply take each channel and each pixel to be a dimension, so a video with $W \times H$ pixels lives in $\mathbb{R}^{3WH}$.  For the 3D mesh features, we use shape histograms \cite{ankerst19993d} after mean centering and RMS normalizing each mesh.  Our shape histograms have 20 radial shells, each with 66 sectors per shell with centers equally distributed across the sphere.  We perform an experiment where we take the 2D features from a face and the 3D features of the same face which has been warped in time.  We perform 10 such warpings and alignments for 9 faces from 6 different expression types.  Figure~\ref{fig:WeizmannBU} shows the aggregated results.  The performance is strikingly worse than the Weizmann dataset, though the normalized IBDTW has about half of the alignment error of other techniques.

\subsection{Non-Euclidean Examples}
\label{sec:noneuclidean}

\begin{figure}
\centering
\includegraphics[width=\columnwidth]{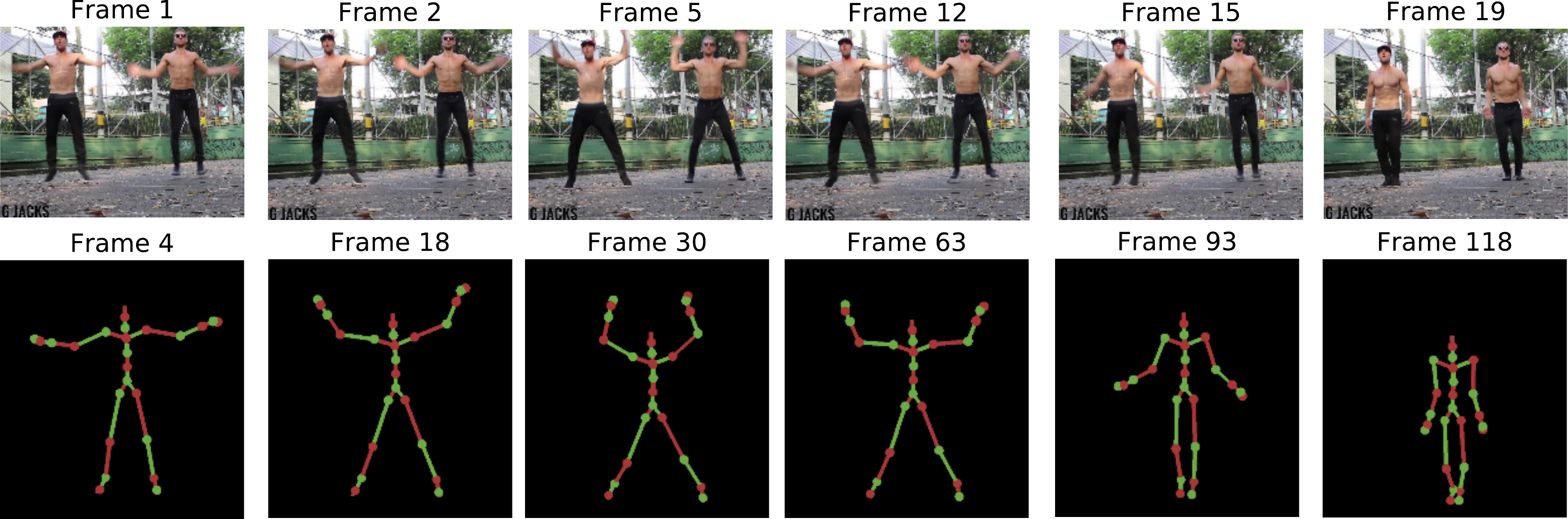}
\caption{Example aligning MOCAP data expressed in the product space of quaternions with videos in raw pixel space.}
\label{fig:MOCAP}
\end{figure}

\begin{figure}
\centering
\includegraphics[width=\columnwidth]{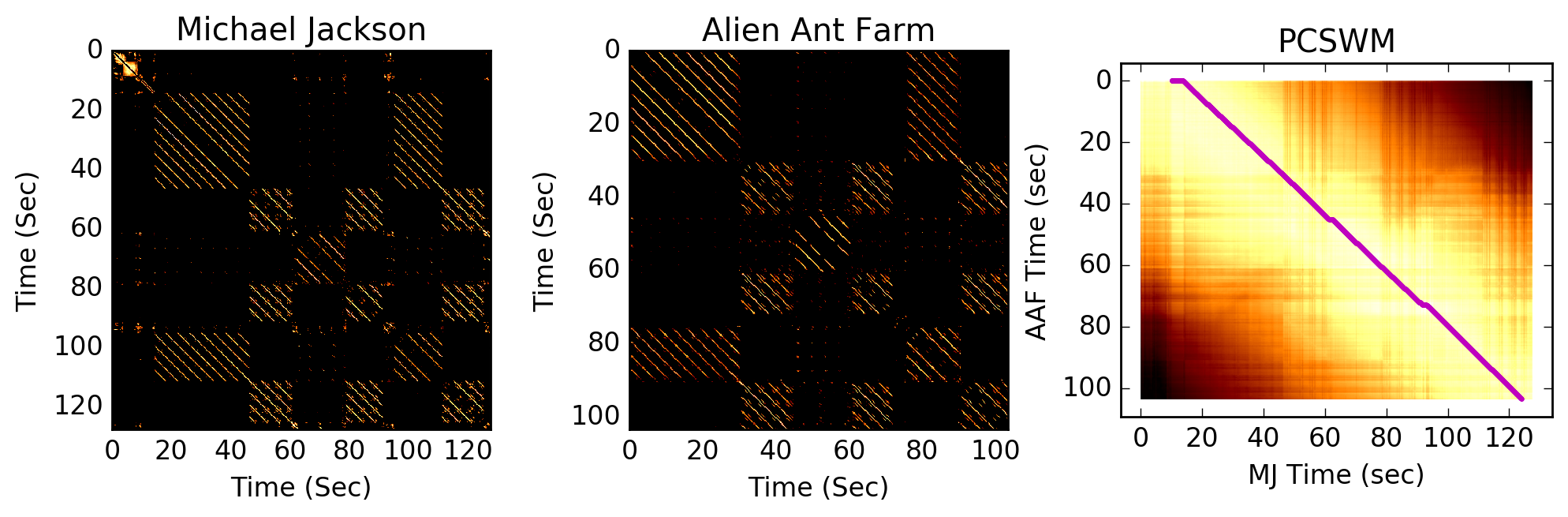}
\caption{Example aligning ``Smooth Criminal'' by Michael Jackson to a faster tempo cover by Alien Ant Farm.  The optimal warping path is superimposed in magenta over the PCSWM.  MJ's version has an intro which is not present in the AAF version, and which is properly skipped by the warping path.}
\label{fig:CoverSong}
\end{figure}

One strength of our algorithms is that they run without modification for features in arbitrary metric spaces.  For example, Figure~\ref{fig:MOCAP} shows IBDTW between motion capture data expressed in a product space of quaternions of $N$ joints with video data expressed as raw pixels.  The product space of quaternions we choose is $\sum_{i=1}^N \cos^{-1}(|q_i \cdot p_i|)$ between two sets of quaternions $(q_1, q_2, ..., q_N)$ and $(p_1, p_2, ..., p_N)$.  For our second example, Figure~\ref{fig:CoverSong} shows IBPTW used to align two ``cover songs'' (two versions of the same song with different instruments / tempos) after fusing note-based and timbral features using similarity network fusion \cite{wang2012unsupervised, wang2014similarity} to learn an improved metric for self-similarity (see \cite{tralieearly} for more info on this process).  Please also refer to supplementary material for video and audio for these examples.

\section{Conclusions}

In this work, we have shown it is possible to synchronize time-ordered point clouds that are spatially transformed without explicitly uncovering the spatial transformation.  As we have shown by our experiments, our algorithms perform excellently when aligning nearly isometric sampled curves, and, with proper normalization, we are competitive with state of the art unsupervised techniques for cross-modal applications.  Furthermore, IBDTW requires no parameters,  making it approachable ``out of the box,'' while, in our experience, CTW and GTW require many parameters that make or break performance.  Finally, we have opened the door for straightforward non-Euclidean time warping, and we hope to see more such applications (\eg time series of graphs).

\section*{Acknowledgments}
Christopher Tralie was partially supported by an NSF Graduate Fellowship NSF under grant DGF-1106401 and an NSF big data grant DKA-1447491.  Paul Bendich is thanked for the suggestion to compare the technique to the Gromov-Hausdorff Distance, and for helpful comments on an initial manuscript.

{\small
\bibliographystyle{ieee}
\bibliography{ssmtw}
}

\end{document}